\documentclass[lettersize,journal]{IEEEtran}
\usepackage{amsmath,amsfonts}
\usepackage{algorithmic}
\usepackage{algorithm}
\usepackage{array}
\usepackage[caption=false,font=normalsize,labelfont=sf,textfont=sf]{subfig}
\usepackage[table]{xcolor}

\usepackage{textcomp}
\usepackage{stfloats}
\usepackage{url}
\usepackage{multirow}
\usepackage{amsthm}
\usepackage{verbatim}
\usepackage{graphicx}
\usepackage{cite}
\usepackage{enumitem} 
\usepackage{booktabs}

\usepackage{bbding}

\newcommand{\green}[1]{\textcolor[rgb]{0,1,0}{#1}}
\newcommand{\blue}[1]{\textcolor[rgb]{0,0,1}{#1}}
\newcommand{\red}[1]{\textcolor[rgb]{1,0,0}{#1}}

\begin{document}

\title{Distractor-aware Event-based Tracking}

\author{Yingkai~Fu, Meng~Li, Wenxi~Liu, Yuanchen~Wang, Jiqing~Zhang, Baocai~Yin, Xiaopeng~Wei, Xin~Yang\textsuperscript{$\dagger$}
\IEEEcompsocitemizethanks{
\IEEEcompsocthanksitem Yingkai Fu, Yuanchen Wang, Jiqing Zhang, Xiaopeng Wei, Xin Yang are with the Key Laboratory of Social Computing and Cognitive Intelligence, Ministry of Education, Dalian University of Technology, Dalian 116024, China (e-mail: yingkai.fu@outlook.com; wangyc0604@mail.dlut.edu.cn; jqz@mail.dlut.edu.cn; xpwei@dlut.edu.cn; xinyang@dlut.edu.cn).
\IEEEcompsocthanksitem Meng~Li is with HiSilicon (Shanghai) Technologies Company Ltd., Shanghai 201799, China (e-mail: limeng8@hisilicon.com).
\IEEEcompsocthanksitem Wenxi~Liu is with the College of Computer and Data Science, Fuzhou University, Fuzhou 350108, China (e-mail: wenxi.liu@hotmail.com).
\IEEEcompsocthanksitem Baocai~Yin is with the Faculty of Information Technology, Beijing University of Technology, Beijing 100124, China (e-mail: ybc@bjut.edu.cn). 
\IEEEcompsocthanksitem \textsuperscript{$\dagger$} Xin~Yang is the corresponding author.

}}
\markboth{IEEE TRANSACTIONS ON IMAGE PROCESSING}%
{Fu \MakeLowercase{\textit{et al.}}: Distractor-aware Event-based Tracking}


\maketitle

\begin{abstract}
Event cameras, or dynamic vision sensors, have recently achieved success from fundamental vision tasks to high-level vision researches. Due to its ability to asynchronously capture light intensity changes, event camera has an inherent advantage to capture moving objects in challenging scenarios including objects under low light, high dynamic range, or fast moving objects. Thus event camera are natural for visual object tracking. However, the current event-based trackers derived from RGB trackers simply modify the input images to event frames and still follow conventional tracking pipeline that mainly focus on object texture for target distinction. As a result, the trackers may not be robust dealing with challenging scenarios such as moving cameras and cluttered foreground. In this paper, we propose a distractor-aware event-based tracker that introduces transformer modules into Siamese network architecture (named DANet). Specifically, our model is mainly composed of a motion-aware network and a target-aware network, which simultaneously exploits both motion cues and object contours from event data, so as to discover motion objects and identify the target object by removing dynamic distractors. Our DANet can be trained in an end-to-end manner without any post-processing and can run at over 80 FPS on a single V100. We conduct comprehensive experiments on two large event tracking datasets to validate the proposed model. We demonstrate that our tracker has superior performance against the state-of-the-art trackers in terms of both accuracy and efficiency.
\end{abstract}


\section{Introduction}

\IEEEPARstart{V}{i}sual object tracking is an important topic in computer vision filed, where the target location is known in the first frame of the video and the algorithm has to identify the target for the rest frames. However, there still remains challenging for the current tracking algorithms and sensors to achieve better accuracy in the environment where the illumination is unstable and the target moves at high speed. On the contrary, the event camera is a novel bionic sensor designed to imitate the imaging mechanism of the human retina. Comparing to the traditional vision sensors like commodity RGB cameras, it 
has high temporal resolution, very high dynamic range, low power consumption and high bandwidth\cite{gallego2020event}. In recent years, event cameras have been employed in more and more areas including object classification \cite{deng2021learning}, object detection \cite{li2022asynchronous}, robotic collision avoidance \cite{sanket2020evdodgenet,falanga2020dynamic}, video reconstruction \cite{scheerlinck2020fast,rebecq2019high,pan2019bringing,cadena2021spade}, SLAM \cite{kim2016real,mueggler2017event,jiao2021comparing,vidal2018ultimate,zhou2021event}, etc.

\def\wdenoising{1 \linewidth}
\begin{figure}[t]
	\setlength{\tabcolsep}{1.2pt}
	\centering
	\begin{tabular}{cc}
	\includegraphics[width=\wdenoising]{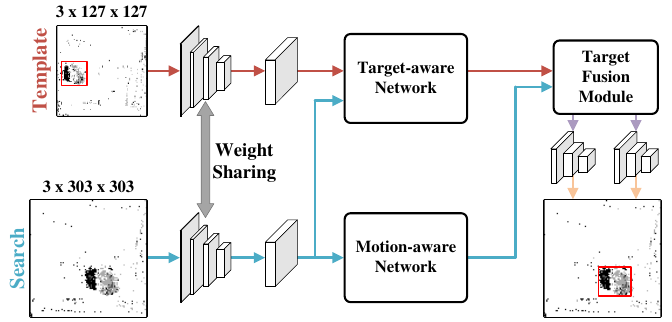}\\	
	\end{tabular}
	
	\caption{The overview of the proposed event-based tracking pipeline. The target-aware network identifies and extracts the target embedding by matching the object edge, the motion-aware network extract moving object features. The target motion fusion module fuses the target embedding and moving objects to locate the moving target. Both the motion and edge information of the event camera can be utilized for precise tracking with the proposed network.}
	\label{fig:teaser}
\end{figure} 

However, despite the obvious advantages of event cameras, designing a visual object tracking algorithm based on event cameras is still challenging. Event cameras can capture changes in light intensity. Typically, they generate more events along the edges of moving objects, where there are larger variations in light intensity, while fewer events occur within the object boundaries where the light intensity changes less. As a result, event cameras cannot extract texture information from the internal of the target effectively, thus most RGB-based trackers that identify and track objects by rich object texture information are not suitable for event-based tracking. Proposing new pipelines for event-based tracking is crucial. ~\cite{barranco2018real,glover2017robust} distinguish between background and moving targets by clustering algorithms, however, the introduced handcrafted strategies are not generalizable across scenarios. Recently, a line of works~\cite{chen2020end, zhang2021object, zhang2022spiking} relies on deep learning to improve the capability of feature extraction, thereby enhancing the discrimination of targets. However, these methods do not explicitly model the relationship between targets and motion. Therefore, when the scene contains distractors such as similar co-moving objects or background noises, the approach is not effective in locating the target.

In this work, we consider both the target distinction and movement information provided by event-based camera for robust tracking, where the target distinction provided by edge information aims to identify the target, and the movement information helps continuously track the object \cite{zheng2022spike}. As shown in Figure \ref{fig:teaser}, we propose a novel event-based object tracking framework (named DANet). In particular, we mainly divide the tracking process in two stages, i.e., \textit{\textbf{discovery}} and \textit{\textbf{identification}}, in which \textit{\textbf{discovery}} aims to find out the potential dynamic objects from current event frame, while the goal of \textit{\textbf{identification}} is to identify the target object with templates. Accordingly, our DANet is mainly composed of two sub-networks: motion-aware network and target-aware network. In specific, the motion-aware network observes the current search frame to distinguish the dynamic objects in the scene. The scene dynamics are extracted via convolutional layers and then passed into transformer-based modules in which the self-correlation of scene features can be computed to discover the significant contrast regions between foreground objects and background. On the other hand, the target-aware network is fed with both the template frame and search frame, and identifies the target object with the help of the template frame. Concretely, the response maps for the template frame and the search frame are extracted through the correlation, then the initial response maps are refined through the transformer-based encoder-decoder module to establish the correlation between target object events and contextual events, resulting in the target embeddings of the target object.Finally, the motion feature from the motion aware network is guided by the output target embedding from the target aware network to generate a confidence map for localizing the target objects. Comprehensive experiments are conducted on public event-based tracking datasets include FE240 dataset \cite{zhang2021object} and VisEvent dataset \cite{wang2021visevent}. Experimental results show our DANet achieves the state-of-the-art performance in accuracy and efficiency, compared with prior approaches.
Overall, the main contributions include:
\begin{itemize}[leftmargin=*]
\item {
We propose a novel event-based tracking model that exploits motion cues and object contours to discover and identify target objects, which facilitates awareness and discrimination in challenging tracking scenarios.}
\item {Our proposed tracker consists of two sub-networks: motion-aware network and target-aware network. Motion-aware network employs transformer-based module to extract scene dynamics, while target-aware network correlates the current frame and search frame to remove distractors.}
\item {The comprehensive experiments on FE240 and VisEvent datasets show that our proposed framework can achieve state-of-the-art performance compared to existing trackers.}
\end{itemize}

The rest of this paper is organized as follows. Section II introduce the fundamentals and related works of vision transformer and the visual object tracking task. In Section III, we explain the details of the proposed DANet. The experimental results, ablations and analysis are provided in Section IV. We conclude the paper in Section V.

\section{Related Work}

\label{sec:related}
\subsection{Vision transformer}
Recently, the vision transformer becomes popular as its ability to capture long range information. It has been applied for many computer vision tasks, like image super-resolution \cite{zhang2018image,yang2020learning,kasem2019spatial}, object detection \cite{carion2020end,sun2021rethinking,fang2021you,9810116} and visual object tracking \cite{chen2021transformer,wang2021transformer,cao2021hift}. 
Transformer are designed to replace the recurrence and convolutions entirely with muti-head self-attention mechanism to capture global information and reallocate the features \cite{chen2021transformer}. The attention mechanism tries to enhance the regions while fading out the non-relevant information by learning relationships between elements of a sequence while keeping their entire composition intact with query, key and value \cite{vaswani2017attention}. In this work, we utilize the transformer architecture to decode target embedding and encode motion cues so as to exploit the contrast difference between dynamic target and background from global perspective for distractor-aware event-based tracking.

\def\wdenoising{1 \linewidth}
\begin{figure}[t]
	\setlength{\tabcolsep}{1.2pt}
	\centering
	\begin{tabular}{cc}
	\includegraphics[width=\wdenoising]{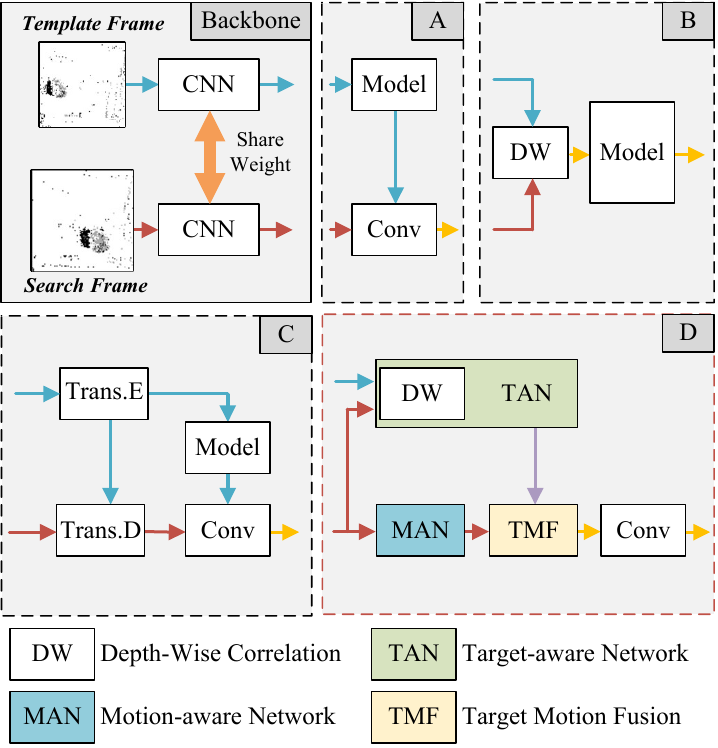}\\	
	\end{tabular}

	\caption{A simplified demonstration of mainly trackers, which can be divided into Siamese-like tracker (A), DCF tracker (B) and transformer tracker (C) \cite{wang2021transformer}. In contrast to the existing trackers. our proposed tracker (D) is designed to make use of both the appearance cue and motion cue from the event camera for robuster event-camera based object tracking.}
	\label{fig:others}
\end{figure} 
\subsection{Visual Object Tracking.}
{Visual object tracking is a central topic of discussion in computer vision, which aims at predicting the target position given the initial state (\textit{i.e.}. the bounding box of the target). The task remains challenging as the appearance of tracking target changes due to light variation, and similar moving target \cite{wang2019fast,smeulders2013visual,yilmaz2006object,li2013survey}.
The mainstream deeplearning-based RGB tracking approaches can be divided into three parts, the discriminative correlation filter (DCF) based tracker that leverage a correlation-based network to compute the similarity between the template and the search region for target tracking \cite{bertinetto2016fully,dai2019visual,li2019gradnet,nam2016learning,zhang2019deeper,8307468,9369121}, the Siamese-like tracker, which learn a model predictor and use it for judge target \cite{danelljan2019atom,bhat2019learning,danelljan2020probabilistic}
and transformer-based tracker \cite{yan2021learning,chen2021transformer,wang2021transformer}. We shows the differences of these pipelines in Figure \ref{fig:others}(A, B, C).
These tracking pipelines work well in the RGB domain as the RGB frame can supply enough texture information for feature matching and object discrimination, however, for scenes like low light, HDR and fast motion, these pipelines can not work efficient as the visual cues are insufficient. To deal with the above nature defects of RGB camera, there are increasing works use event camera for visual tracking to settle scenarios with challenging light variation. In traditional-based methods, Mondal \textit{et al.} \cite{mondal2021moving} presented an unsupervised graph spectral clustering technique for moving object location in event-based data. Barranco \textit{et al.} \cite{barranco2018real} proposed a real-time mean-shift clustering algorithm for event-based multi-object tracking. In deep learning-based methods, Chen \textit{et al.} \cite{chen2020end}  proposed a novel event representation to encode the spatio-temporal information and designed a 5-DoF object-level motion model to regress object motions. In multi-domain event-based tracking tasks, Zhang \textit{et al.}\cite{zhang2021object,zhang2023frame} proposed attention scheme and multi-modality alignment to fuse the frame domain and event domain for single object tracking, Zhu \textit{et al.} \cite{zhu2023cross} proposed a mask modeling strategy on pre-trained vision transformer to enforce the interaction from RGB and Event modalities, Wang \textit{et al.} \cite{tang2022revisiting} proposed a unified trmer backbone network for color-event tracking. However, these event tracker are designed to leverage the edge cues, therefore, when the camera is moving, these trackers failed to distinguish target from background. Actually, an event-based camera does not generate semantic information, it offers rich object motion information as moving objects can typically generate more events due to the large contrast changes at the object edge \cite{zhang2021object,gallego2020event}. In this paper, we exploit the motion cues for robust tracking.}

\section{Our Proposed Method}

\subsection{Framework Overview}

As illustrated in Figure \ref{fig:teaser}, to facilitate the \textit{\textbf{discovery}} and \textit{\textbf{identification}} ability of the proposed pipeline, our framework is mainly composed of feature extraction network, target-aware network, motion-aware network, target fusion module, as well as object localization regressor. In particular, the feature extraction module first extracts the features for the template and search frames. Then, the features of the search frame are fed into the motion-aware network that extracts scene dynamics, while those of template and search frames are both sent into the target-aware network so as to remove dynamic distractors and focus on the target object. Finally, their outcomes are fused in the target fusion module and employed to regress the target location.
We will describe the details of each component in the remainder of this section. The details of the target-aware network, the motion-aware network and target fusion module are visualized in Figure \ref{fig:overview}.

\def\wdenoising{1.0\linewidth}
\begin{figure*}[htb]
    \begin{center}
    \includegraphics[width=\wdenoising]{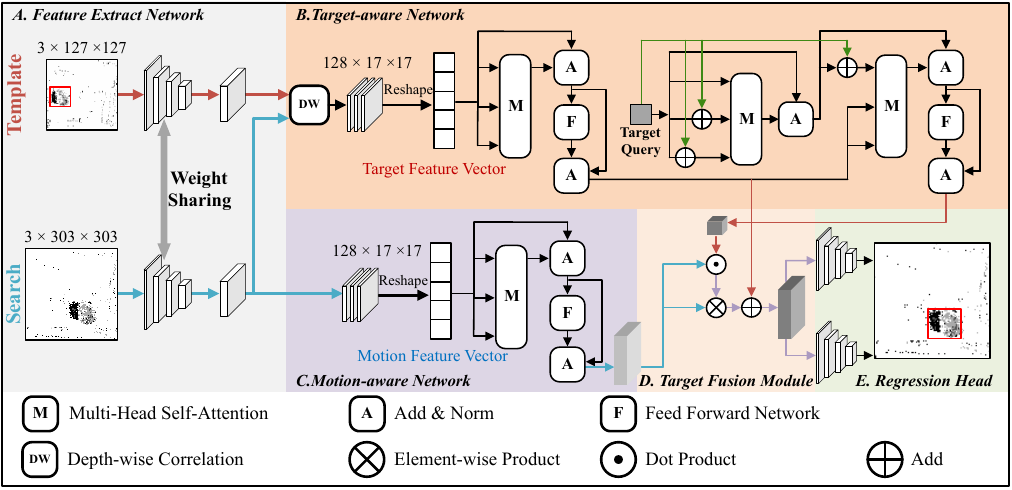}

    \end{center}
    \caption{The details of five main parts proposed in our tracker, namely Feature Extract Network, Target-aware Network, Motion-aware Network,Target Fusion Module and Regression Head. The search feature extract from search frame come to both target-aware network and motion-aware network to provide target features and movement features.}
    \label{fig:overview}
\end{figure*}
\subsection{Event Representation}
\label{sec:event_rep}
Different from conventional frame-based cameras, event-based cameras produce asynchronous data streams at high speed, when any pixel detects a log scale intensity change that exceeds a threshold $C$, as:
\begin{equation}
L(x, y, t)-L(x, y, t-\Delta t) \geq p C,
\end{equation}
where $L$ denotes log scale light intensity, $C$ is the contrast threshold determined by the hardware, and $p\in \{-1,+1\}$ is the event polarity depending on the change in brightness. Mathematically, a set of events can be described as:
\begin{equation}
E=\left\{e_{k}\right\}_{k=1}^{N}=\left\{\left[x_{k}, y_{k}, t_{k}, p_{k}\right]\right\}_{k=1}^{N},
\end{equation}
where $k$ is the index of each event, $x_{k}$ and $y_{k}$ represent the pixel location, $t_k$ denotes the timestamps of the event.

\def\wdenoising{1\linewidth}
\begin{figure}[t]
    \setlength{\tabcolsep}{0.8pt}
    \centering
        
    \begin{tabular}{cc}
        
    \includegraphics[width=\wdenoising]{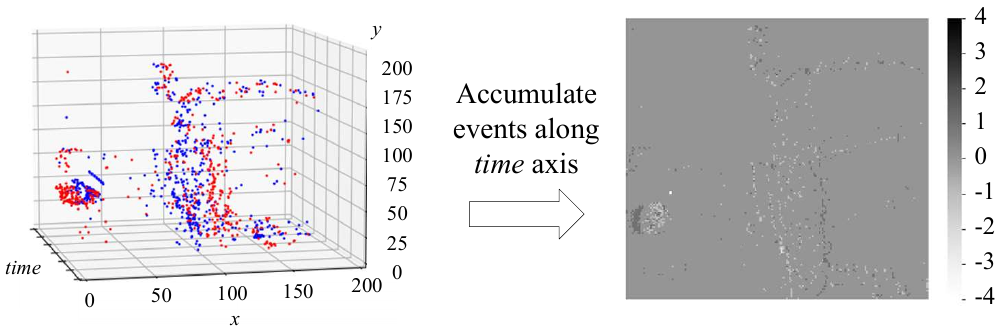}
            
    \end{tabular}
    \caption{The event steam is accumulated along $t$ axis to generate an event frame, the moving part can be emphasized and some background noises are filtered out.}
\label{fig:stack}
\end{figure} 
Yet, the asynchronous events cannot be processed directly due to their sparse nature, and thus algorithms designed for conventional frames are not applicable. To address this issue, for each pixel, we propose to sum up the polarity of sparse event data within a fixed time interval to obtain event frames, The event stream and the stacked event frame are illustrate in Figure \ref{fig:stack}, the stacking process can be formulated as:
\begin{equation}
F({x,y,\Delta t})=\sum_{k=i}^{N}p_k(x,y),
\end{equation}
There are several benefits for aggregating event data in this way. First, the original sparse event stream can be enhanced, while the event frames can still reflect object contours as the events caused by little contrast change will be filtered out. 
Second, the aggregated offset values in event frames can reflect the extent of object motion, which can be used to better identify different moving objects. Besides, the event aggregation can be efficiently computed in parallel and it is easy to integrate it into existing event camera development kits without compromising real-time data processing and tracking capabilities.

\subsection{Event Feature Extraction}
Based on Siamese network architecture, our event-based feature extraction module is fed with the template frame, $z \in \mathbb{R}^{H_z \times W_z}$, and the search template, $x \in \mathbb{R}^{H_x \times W_x}$. Both frames are passed into a shared network based on ResNet-18 \cite{he2016deep} for extracting event features. Thus, we can obtain the event features of $z$, i.e., $f(z) \in \mathbb{R}^{C \times \frac{Hz}{s} \times \frac{Wz}{s} } $, and those of $x$, $f(x) \in \mathbb{R}^{C \times \frac{Hx}{s} \times \frac{Wx}{s} }$, where $s$ is downsampling ratio. These features are then input to the proposed Target-aware Network and Motion-aware Network for extracting the similar target and moving objects.

\subsection{Target-aware Network (TAN)}
The detail of the target-aware network is illustrated in Figure \ref{fig:overview}(B), the TAN aims to match the given tracking target at the template frame with the search frame. To be specific, firstly, the features of the template frame $f(z)$ and the search frame $f(x)$ are separately passed to two convolution blocks with 3$\times$3 kernel size. Then, the template feature embeddings are employed as a kernel to compute convolution on the feature embeddings of the search frame to produce an initial target response map $R$, which implies the confidence of object localization. Formally, we have:
\begin{align}
    R = \psi(z, x)=\phi(f(z)) \star \varphi(f(x)),
\end{align}
where $f(\cdot)$ refers to feature extraction network. $\phi(\cdot)$ and $\varphi(\cdot)$ are two convolutional layers. $\star$ represents the depth-wise cross-correlation operator, which enables to imply the coarse correlation between the contour of target object and the events within search region.
To refine the initial target response map while removing distractors, we present transformer-based modules to strengthen the event information from a global perspective. Firstly, we review multi-head attention mechanism, as:
\begin{align}
& {MultiHead}({Q}, {K}, {V})= {Concat}( {H}_{1}, ,  {H}_{n}) {W}^{o} \\
& {H}_{i}= {Attention}({QW}_{i}^{Q}, {KW}_{i}^{K}, {VW}_{i}^{V}), \\
& {Attention}({Q}, {K}, {V})=\operatorname{softmax}({Q K}^{\top}/{\sqrt{d_{k}}}) {V},
\end{align}
where ${W}_{i}^{Q} \in \mathbb{R}^{d_m \times d_k} $, $\text{W}_{i}^{K} \in \mathbb{R}^{d_m \times d_k} $,$\text{W}_{i}^{V} \in \mathbb{R}^{d_m \times d_v} $, and $\text{W}^{O} \in \mathbb{R}^{n_hd_m \times d_m} $ denote model parameters.

In TAN, we use Transformer to enable target embedding. To do so, the initial target response map is first passed through a bottleneck layer to squeeze the features by reducing the channel number from $C$ to $d$, and then flatten the feature maps along the spatial dimension to form a feature vector. Next, the correlation $R$ is fed to the encoder blocks to utilize the contextual information. The encoder block consists of multi-head self-attention modules and feed forward network with two linear transformation with a ReLU in between to enhance the fitting ability of the model. The output $R^\prime$ of an encoder block can be described as:
\begin{align}
&R^\prime= \mathcal{F}_e+ {FFN}(\mathcal{F}_e), \\
&\mathcal{F}_e= R+  {MultiHead}(R, R,R ),\\
&{FFN}({x})=\max ({0}, {x} {W}_{1}+{b}_{1}) {W}_{2}+{b}_{2},
\end{align}
Where ${W}$ and $ {b}$ are weight matrix and basis vectors, respectively. We further exploit a decoder module to obtain the embeddings of the target object, the inputs for a decoder module is a learnable target object query and the output from encoder. The decoder module can be expressed as below:
\begin{align}
&T = \mathcal{F}_d+ {FFN}(\mathcal{F}_d),\\
&\mathcal{F}_d= A+  {MultiHead}(A+TQ, R^\prime,R^\prime),\\
&A = TQ +  {MultiHead}(TQ+TQ,TQ+TQ,TQ),
\end{align}
where $TQ$ is a target query. We only input one single query into the decoder to predict one bounding box of the target object as in \cite{yan2021learning}.

\subsection{Motion-aware Network (MAN)}
{As introduced in Section \ref{sec:event_rep}, by accumulating the events within a short time interval into an event frame and adding up the polarity for each pixel, the ability to capture moving objects can be preserved. Therefore, event frames can be used for discovering all the moving objects within the scene, since moving objects generally trigger more events compared to the background.}

{To accomplish this, the extracted features of the search frame are firstly passed to a $3\times3$ convolutional layer with stride 2 to compress and reshape the features to $M$, it is further sent into the motion-aware network (see Figure \ref{fig:overview}(C)). The structure of the motion-aware network consists of the transformer encoder module that is identical to the target-aware network. The motion-aware network works as:} 
\begin{align}
&M^\prime= \mathcal{M}_e+ {FFN}(\mathcal{M}_e), \\
&\mathcal{M}_e= M+  {MultiHead}(M, M, M).
\end{align}

\subsection{Target Fusion Module (TFM)}
In the final stage, we propose to fuse the target embeddings $T$ and the motion feature maps $M^\prime$ to generate the refined heat map for regressing moving target. Specifically, as shown in Figure \ref{fig:overview}(D), we take target embedding from the output of target-aware network, then compute the similarity between the target features and the output motion features from the motion-aware network. Next, we apply the similarity scores to the motion features with an element-wisely multiply to enhance the regions for the target objects and weaken other moving objects. To effectively utilize the output of the encoder in TAN, we introduce a residual connection after multiplication to fuse two outcomes, the output feature maps are then used to regress object boundaries.

\subsection{Regression Head}
The regression head aims to locate the target with the enhanced feature maps. The enhanced feature is passed through two fully convolution networks with ReLU activation function to formulate the features for predicting object center and the scale. Finally, we apply the soft-argmax function to extract coordinates via calculating a differentiable expectation \cite{nibali2018numerical}. Formally, we have:
\begin{align}
c_{x}(h)=\sum_{i=1}^{W} \sum_{j=1}^{H}  \text{softmax} \left(h_{i, j} \right) w_{i, j, x},\\
c_{y}(h)=\sum_{i=1}^{W} \sum_{j=1}^{H}  \text{softmax} \left(h_{i, j} \right) w_{i, j, y},\\
w_{i, j, x}=\frac{i}{W} \times s, w_{i, j, y}=\frac{j}{H}  \times s,
\end{align}
{where $h_{i,j}$ denotes the output value at location $(i,j)$ on feature maps $h$, $s$ denotes the total stride that down sample from the original images size to the size of feature maps. }

\subsection{Loss Function}
DANet does not require any postprocessing, therefore, it can be trained in an end-to-end fashion. We use the combination of $L_1$ Loss and the generalized IoU loss \cite{rezatofighi2019generalized} as in \cite{carion2020end} to regulate the training to predict target location. The loss function is formulated as:
\begin{equation}
   L=\lambda_{\text {iou}} L_{\text {iou}}\left(b_{i}, \hat{b}_{i}\right)+\lambda_{L_{1}} L_{1}\left(b_{i}, \hat{b}_{i}\right) ,
\end{equation}
where $b_i$ and $\hat{b}_{i}$ represent the ground-truth and the predicted box. $\lambda_{iou}$ and $\lambda_{L_1}$ are balancing weights.

\def\wdenoising{1\linewidth}
\begin{figure*}[!thb]
    \begin{center}
    \includegraphics[width=\wdenoising]{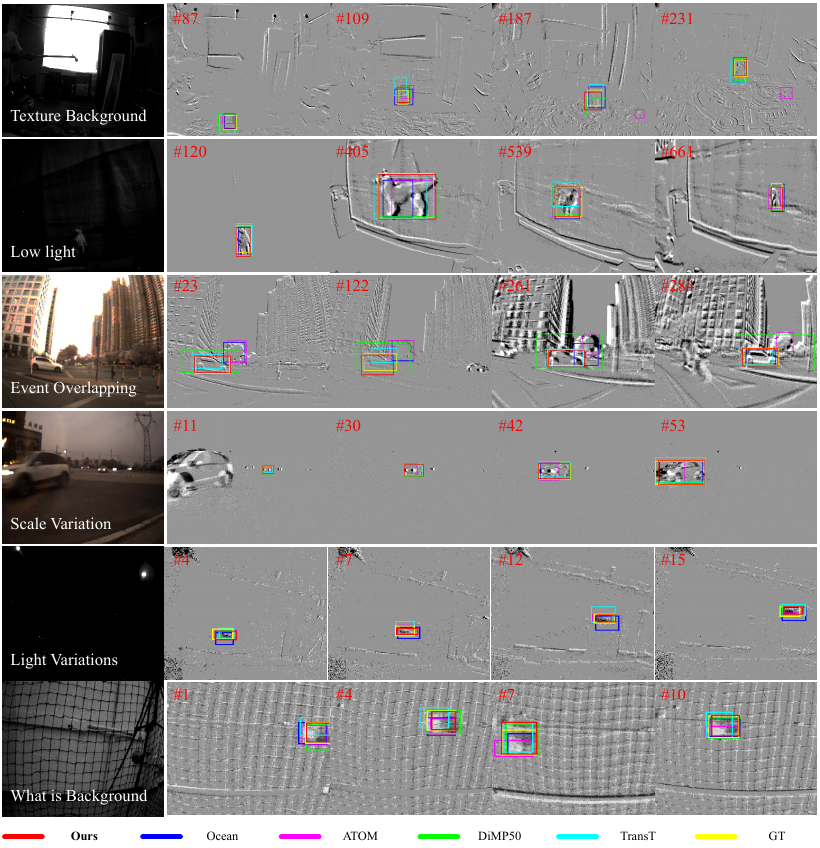}
    \end{center}
    \caption{Visual comparison of our proposed tacker against other state-of-the-art trackers on FE240 \cite{zhang2021object} dataset (The first two rows), VisEvent \cite{wang2021visevent} dataset (row 3-4) and EED dataset \cite{mitrokhin2018event} (row 5-6). The images in first column are only used for visualization.}
\label{fig:sup_effect}
\end{figure*}
\section{Experimental Results}
\subsection{Datasets and Implementation Details}
\paragraph{Datasets}
We use three event camera tracking datasets to validate the proposed tracker, namely FE240 dataset \cite{zhang2021object}, VisEvent dataset \cite{wang2021visevent} and EED dataset \cite{mitrokhin2018event}.

{\bf{FE240 dataset}} is a large indoor event camera tracking dataset that contains various degraded scenarios, such as high dynamic range, low light, motion blur, and fast motion. It features 82 long sequences. 
All of them are captured by DAVIS346 event camera, which simultaneously capture events and calibrated frames. The ground-truth bounding boxes of the targets are annotated by a motion capture system at 240Hz. We use 57 sequences to train and the rest 25 sequences for testing.

{\bf{VisEvent dataset}} is a large-scale neuromorphic tracking dataset collected from real world by a DAVIS346 color event camera. The VisEvent dataset includes challenging scenarios in various condition, \textit{e.g.}, low illumination, high speed, clutter background, and slow motion. The sequences from this dataset contain RGB frames and events at the resolution of $346 \times 260$. The annotation for each frame is fulfilled by a professional label company at about 25 FPS. We exclude 97 training sequences and 23 test sequences from the original dataset due to wrong annotations or missing events, so there are 403 training sequences and 297 testing sequences used in our experiments.

{\bf{EED dataset}} is a validate only event-based tracking dataset collected by a DAVIS event camera under extreme condition such as strobe light, occlusion and cluster background. As the annotation for EED dataset is not precise, we manually label the dataset and generate event frame from the provided event files.

\def\wdenoising{0.5\linewidth}
\def\hdenoising{1.5in}
\begin{figure}[t]
	\setlength{\tabcolsep}{1.2pt}
	\centering
	
	\begin{tabular}{cc}
	
	\includegraphics[width=\wdenoising, height=\hdenoising]{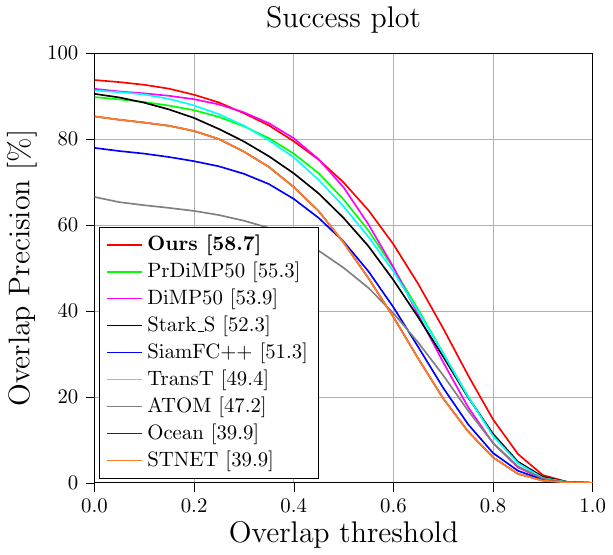}  & 
	\includegraphics[width=\wdenoising, height=\hdenoising]{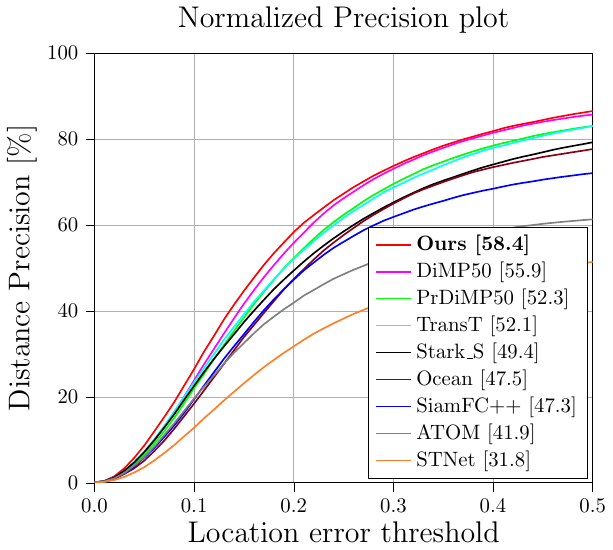}	 \\
		
	\end{tabular}
	\caption{Success plot and normalized precision plot on FE240 dataset \cite{zhang2021object}. In the legend, the area-under-curve (AUC) and normalized precision (P$_{Norm}$) are reported in the left and right figures, respectively. Our proposed method achieve the best performance in both two metric. }
	\label{fig:statistics}
\end{figure} 
\def\wdenoising{0.5\linewidth}
\def\hdenoising{1.5in}
\begin{figure}[t]
	\setlength{\tabcolsep}{1.2pt}
	\centering
	
	\begin{tabular}{cc}
	
		\includegraphics[width=\wdenoising, height=\hdenoising]{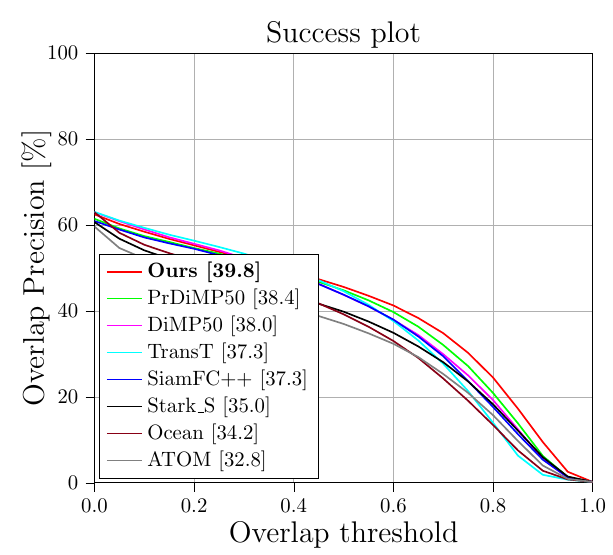} & 
		 \includegraphics[width=\wdenoising, height=\hdenoising]{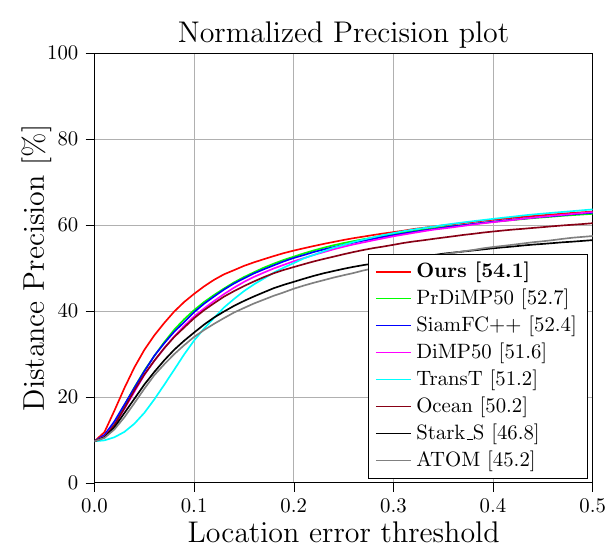} \\
		
	\end{tabular}
	\caption{Success plot and normalized precision plot on VisEvent dataset \cite{wang2021visevent}. Our method has competitive performance compared to other trackers.}
	\label{fig:eed_plot}
\end{figure} 
\paragraph{Metrics}
We follow the experimental protocols of visual tracking as one-pass evaluation (OPE) \cite{wu2013online} and report the following metrics to show the performance of the proposed method, \textit{i.e.}, Area Under Curve (AUC), Overlap Precision (OP), Precision, and Normalized Precision, respectively. We also show the success plot and normalized precision plot for a comprehensive comparison of different thresholds.  

{\bf{Success plot}} shows the percentage of frames with the overlapping rate $ S \geq t_0$  throughout all the thresholds $t_0 \in \left[  0,1 \right] $. The overlapping rate is defined as:
\begin{equation}
S(a, b)=\frac{\operatorname{Area}(a) \cap \operatorname{Area}(b)}{\operatorname{Area}(a) \cup \operatorname{Area}(b)},
\end{equation}
where $\operatorname{Area}(a)$ denotes the bounding box predicted by the tracker, and the $\operatorname{Area}(b) $ means the bounding box given by the ground-truth. The AUC of each success plot serves as the second measure to rank the trackers. The success curve offers a continuous measurement of tracking results ranging from robustness (lower overlap rate but more tracked frames) to accuracy (higher overlap rate) \cite{huang2019got}. The overlap precision metric (OP$_T$) is defined as the percentage of frames where bounding box IoU overlap larger than a threshold $T$.
 
{\bf{Precision plot}} measures the distance between the center of predicted bounding box and that of the ground-truth bounding box. However, the target size and image resolution have large discrepancies for different frames and videos, which will heavily influence the precision metric. Therefore, we only use normalized precision plot for comprehensive comparison for each threshold. The precision rate with the threshold 20 and the normalized precision rate with the threshold 0.2 are used to rank the trackers.

\paragraph{Implementation Details}
The inputs for our tracker are a pair of frames, \textit{i.e.}, the template frame and search frame which are cropped and resized to $127 \times 127$ and $303 \times 303$, respectively. The feature extractor backbone ResNet18 \cite{he2016deep} is pretrained on ImageNet \cite{imagenet_cvpr09}. The following convolutional blocks consist of three stacked Conv-BN-ReLU layers to reduce the feature size and produce the feature maps of $128 \times 17 \times 17$. The head number of multi-head attention is set to 4 with the width 128. In addition, TAN consists three encoders and three decoders, while MAN consists of two encoder blocks. TAN takes an additional target query as input. The feed-forward network in both TAN and MAN has a hidden dimension of 2048 with a dropout ratio of 0.1. The FCN layers in the object localization regressor are five stacked Conv-BN-ReLU layers with the output channel as 128, 64, 32, 16 and 1, respectively, which produce two feature maps that are separately used to produce the center coordinates and the shape of target object. We set $\lambda_{iou}$ to 1 and $\lambda_{L_1}$ to 5 in the loss function.

\paragraph{Training and Network Inference}
{Our tracker is implemented using Python 3.8 and PyTorch 1.9.1 \cite{paszke2019pytorch}. The experiments are conducted on a server with Intel(R) Xeon(R) Gold 6240 CPU @ 2.60GHz, and an NVIDIA V100 32GB GPU with CUDA 11.1. At the training stage, it takes a warm up \cite{devlin2018bert} epoch with the learning rate linearly increased from $10^{-5}$ to $8 \times 10^{-2} $, and then uses a cosine annealing learning rate schedule for the rest 19 epochs. We choose stochastic gradient descent as the optimizer. The batch size is set to 128. For each epoch, $3 \times 10^5$ images pairs are sampled randomly. 
During inference, given the first event frame and the corresponding bounding box, the network initializes the template branch and calculates the template features. The input search frames are cropped around the latest predicted location. Then, the Siamese feature extractor extracts features for the search event frame. After that, the search features and the template features are sent to the succeeding modules to predict a bounding boxes, which represent the normalized location relative to the previous prediction location. Finally, a simple transformation is applied to restore the box scale that fit the original image size. }

\subsection{Comparison with SOTA Trackers}
To verify the efficiency and precision of our proposed DANet, we choose 8 SOTA trackers to be compared with, namely SiamFC++ \cite{xu2020siamfc++}, ATOM \cite{danelljan2019atom}, DiMP50 \cite{bhat2019learning}, PrDiMP50 \cite{danelljan2020probabilistic}, STARKS \cite{yan2021learning}, TransT \cite{chen2021transformer}, Ocean \cite{zhang2020ocean} and STNet\cite{zhang2022spiking}. For a fair comparison, we use their publicly available codes and then retrained these trackers with their default parameters. To validate the performance of each tracker, we report AUC, OP50, OP75, Precision, and Normalized Precision to demonstrate the performance of the tracker. The threshold of Precision and Normalized Precision is set to 20 pixels and 0.2, respectively.

\def\wdenoising{0.5\linewidth}
\def\hdenoising{1.5in}
\begin{figure}[t]
	\setlength{\tabcolsep}{1.2pt}
	\centering
	
	\begin{tabular}{cc}
	
		\includegraphics[width=\wdenoising, height=\hdenoising]{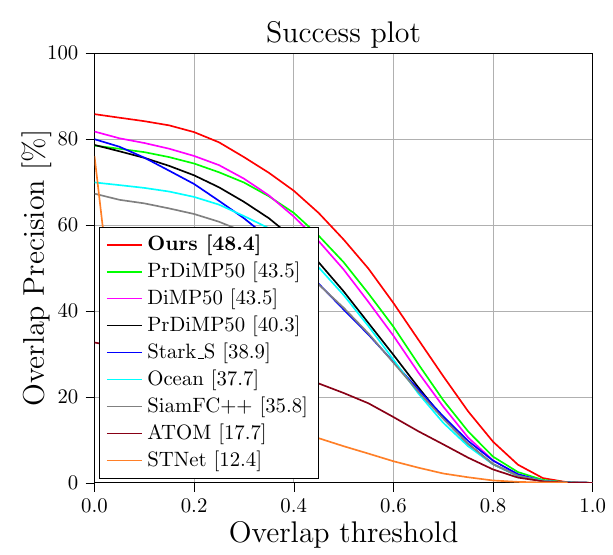} & 
		\includegraphics[width=\wdenoising, height=\hdenoising]{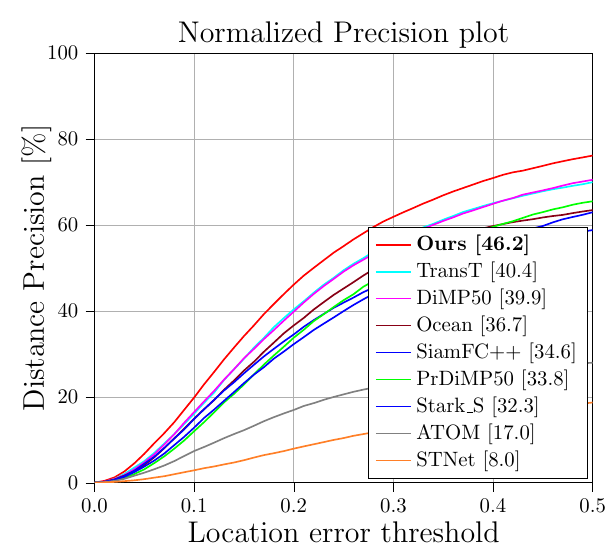} \\
		
	\end{tabular}
	\vspace{-0.3cm}
	\caption{Success plot and normalized precision plot of camera motion sequences on FE240 dataset \cite{zhang2021object}, which demonstrates the proposed tracker has superior tracking performance compared to other algorithms in camera motion scenarios.}
	\label{fig:motion_plot}
\end{figure} 
\def\wdenoising{0.5\linewidth}
\def\hdenoising{1.5in}
\begin{figure}[t]
	\setlength{\tabcolsep}{1.2pt}
	\centering
	
	\begin{tabular}{cc}
	
		\includegraphics[width=\wdenoising, height=\hdenoising]{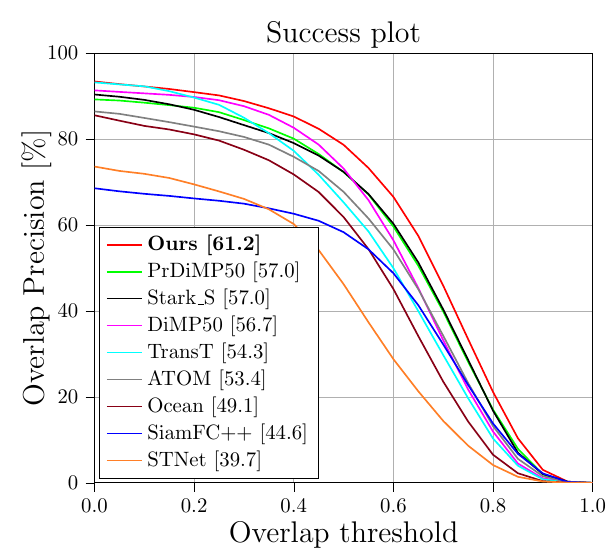} & 
		 \includegraphics[width=\wdenoising, height=\hdenoising]{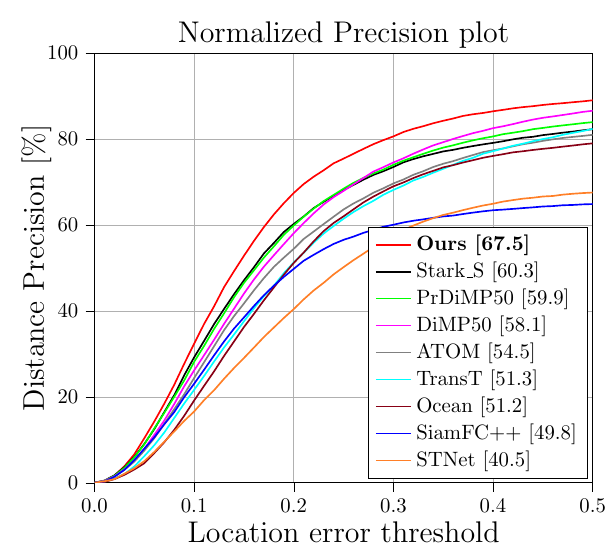} \\
		
	\end{tabular}
	\caption{Success plot and normalized precision plot of similar object sequences on FE240 dataset \cite{zhang2021object}. The proposed tracker demonstrates superior tracking performance compared to other algorithms in similar object scenarios.}
	\label{fig:similar_plot}
\end{figure} 
\setlength{\tabcolsep}{5pt}
\begin{table*}[htb]
	\caption{State-of-the-art comparison on FE240 dataset and VisEvent dataset in terms of AUC, Precision and Normalized Precision. The first, second, and third best results are marked in \red{\textbf{red}}, \green{\textbf{green}}, and \blue{\textbf{blue}}, respectively. our proposed method achieves the best performance on these two dataset under five standard comparing metrics.}
	\centering
	\scalebox{0.85}{
	\begin{tabular}{c|c|ccccccccc}
		\toprule
		\multirow{2}{*}{Dataset} &
		\multirow{2}{*}{Metrics} & \multicolumn{1}{c}{STNet \cite{zhang2022spiking} } & \multicolumn{1}{c}{OCEAN\cite{zhang2020ocean} } &\multicolumn{1}{c}{ATOM\cite{danelljan2019atom}}& \multicolumn{1}{c}{TransT\cite{chen2021transformer}}
		& \multicolumn{1}{c}{Stark\_S\cite{yan2021learning}} & \multicolumn{1}{c}{ SiamFC++\cite{xu2020siamfc++}}
		 & \multicolumn{1}{c}{DiMP50\cite{bhat2019learning}} & \multicolumn{1}{c}{ PrDiMP50\cite{danelljan2020probabilistic}}   & \multicolumn{1}{c}{DANet} \\
		\cmidrule{3-11} 
      	& \multicolumn{1}{c|}{} & CVPR'22 &ECCV'20  & CVPR'19	&CVPR'21  & ICCV'21  &AAAI'20    & ICCV'19  & CVPR'20   &   Ours \\ 
		
		\midrule
		\multirow{5}{*}{FE240\cite{zhang2021object}} 
		&Success(AUC)$(\%)\uparrow$  &31.51 & 47.28 & 40.07 & \textcolor{blue}{53.46} & 51.79 & 45.23 & \textcolor{green}{54.49} & 53.43 & \textcolor{red}{\textbf{56.86}} \\

        &OP50.$(\%)\uparrow$  &36.07 & 55.93 & 50.08 & 64.35 & 61.68 & 56.17 & \textcolor{green}{68.79} & \textcolor{blue}{65.98} & \textcolor{red}{\textbf{69.94}} \\

        &OP75.$(\%)\uparrow$  &6.92  & 11.99 & 16.80  & \textcolor{blue}{20.06} & 19.88 & 13.64 & 17.63 & \textcolor{green}{20.13} & \textcolor{red}{\textbf{25.03}} \\

        &Precision$(\%)\uparrow$  &52.46 & 80.96 & 61.85 & \textcolor{blue}{86.13} & 83.9  & 74.23 & \textcolor{green}{87.55} & 85.14 & \textcolor{red}{\textbf{89.19}} \\

        &Norm. Prec.$(\%)\uparrow$  &31.78 & 47.47 & 41.93 & 52.11 & 49.43 & 47.29 & \textcolor{green}{55.88} & \textcolor{blue}{52.28} & \textcolor{red}{\textbf{58.37}} \\

        \midrule
		\multirow{5}{*}{VisEvent\cite{wang2021visevent} } 
		&Success(AUC)$(\%)\uparrow$ & &34.18 & 32.82 & 37.30  & 34.97 & 37.30  & \textcolor{blue}{38.03} & \textcolor{green}{38.35} & \textcolor{red}{\textbf{39.76}} \\

		&OP50.$(\%)\uparrow$ & &39.25 & 37.01 & \textcolor{blue}{44.63} & 39.86 & 43.81 & 43.80  & \textcolor{green}{44.84} & \textcolor{red}{\textbf{45.62}} \\

		&OP75.$(\%)\uparrow$ & &19.12 & 20.96 & 21.37 & 23.61 & 23.71 & \textcolor{blue}{25.02} & \textcolor{green}{27.18} & \textcolor{red}{\textbf{30.27}} \\

		&Precision$(\%)\uparrow$ & &51.39 & 46.48 & \textcolor{red}{\textbf{54.82}} & 48.19 & 53.08 & \textcolor{blue}{54.08} & 53.93 &\textcolor{green}{54.54}\\

		&Norm Prec.$(\%)\uparrow$ & &50.24 & 45.2  & 51.15 & 46.80  & \textcolor{blue}{52.36} & 51.58 & \textcolor{green}{52.66} & \textcolor{red}{\textbf{54.07}} \\

        \midrule
		\multirow{5}{*}{EED\cite{mitrokhin2018event}} 
		&Success(AUC)$(\%)\uparrow$ &35.50 &\textcolor{blue}{53.45} & 48.36 & 48.98 & 42.63 & 42.81 & \textcolor{green}{53.60}  & 48.62 & \textcolor{red}{\textbf{55.69}} \\
		&OP50.$(\%)\uparrow$ & 39.37 &54.56 & 48.21 & 53.4  & 42.72 & \textcolor{blue}{55.73} & \textcolor{green}{65.64} & 46.76 & \textcolor{red}{\textbf{65.94}}  \\
		&OP75.$(\%)\uparrow$ & 10.79 & 12.36 & \textcolor{blue}{12.97} & 12.09 & 8.41  & 10.79 & \textcolor{green}{19.94} & 10.44 & \textcolor{red}{\textbf{23.43}}  \\
		&Precision$(\%)\uparrow$ & 62.70 &\textcolor{red}{\textbf{100}}   & 85.58 & \textcolor{blue}{88.70}  & 75.47 & 69.47 & 86.78 & \textcolor{green}{91.22} & 86.78 \\
		&Norm Prec.$(\%)\uparrow$ & 31.25 &41.29 & \textcolor{green}{58.50}  & 45.46 & 36.59 & 38.83 & \textcolor{red}{\textbf{58.53}} & \textcolor{blue}{57.03} & 50.56 \\
  
		\bottomrule
	\end{tabular}
	}

	\label{tab:sota}
\end{table*}

\paragraph{Comparison on FE240 Dataset}
As shown in Table \ref{tab:sota}, we can see that our proposed offline tracker surpasses other state-of-the-art trackers on all five evaluation metrics. In particular, our tracker obtains a performance with 56.86\%, 89.19\% and 58.37\% in terms of AUC, Precision, and Normalized Precision, respectively. Note that DANet excels DiMP50 \cite{danelljan2020probabilistic} and PrDiMP50 \cite{danelljan2020probabilistic} by a large margin, which is the state-of-the-art online tracker with ResNet-50 \cite{he2016deep} as backbone. We also report the success plot and normalized precision plot in Figure \ref{fig:statistics}.
Additionally, we analyze the results on camera motion sequences and multiple object motion sequences. The results are displayed in Figure\ref{fig:scenario}. When the camera is moving or multiple objects are moving together, the events from the background or other objects may severely disturb the competitor. However, our tracker still achieves 48.69\% and 61.94\% for these two kinds of scenarios in terms of AUC. For scenes where illuminations are challenging, \textit{e.g.}, HDR, low light and strobe light, our tracker still achieve good results. The visualization of the severe motion and strobe light scenes can be observed on the first and third row in Figure \ref{fig:effect}. Although the events from the target are overlapped with background events, our tracker keeps focusing on the moving target. More visual examples are shown in Figure \ref{fig:sup_effect} row 1-2, which indicates the superior performance of the proposed tracker in challenging scenes.

\def\wdenoising{1\linewidth}
\begin{figure}[t]
	\setlength{\tabcolsep}{0.8pt}
	\centering
	\scalebox{0.97}{
	\begin{tabular}{cc}
	
	\includegraphics[width=\wdenoising]{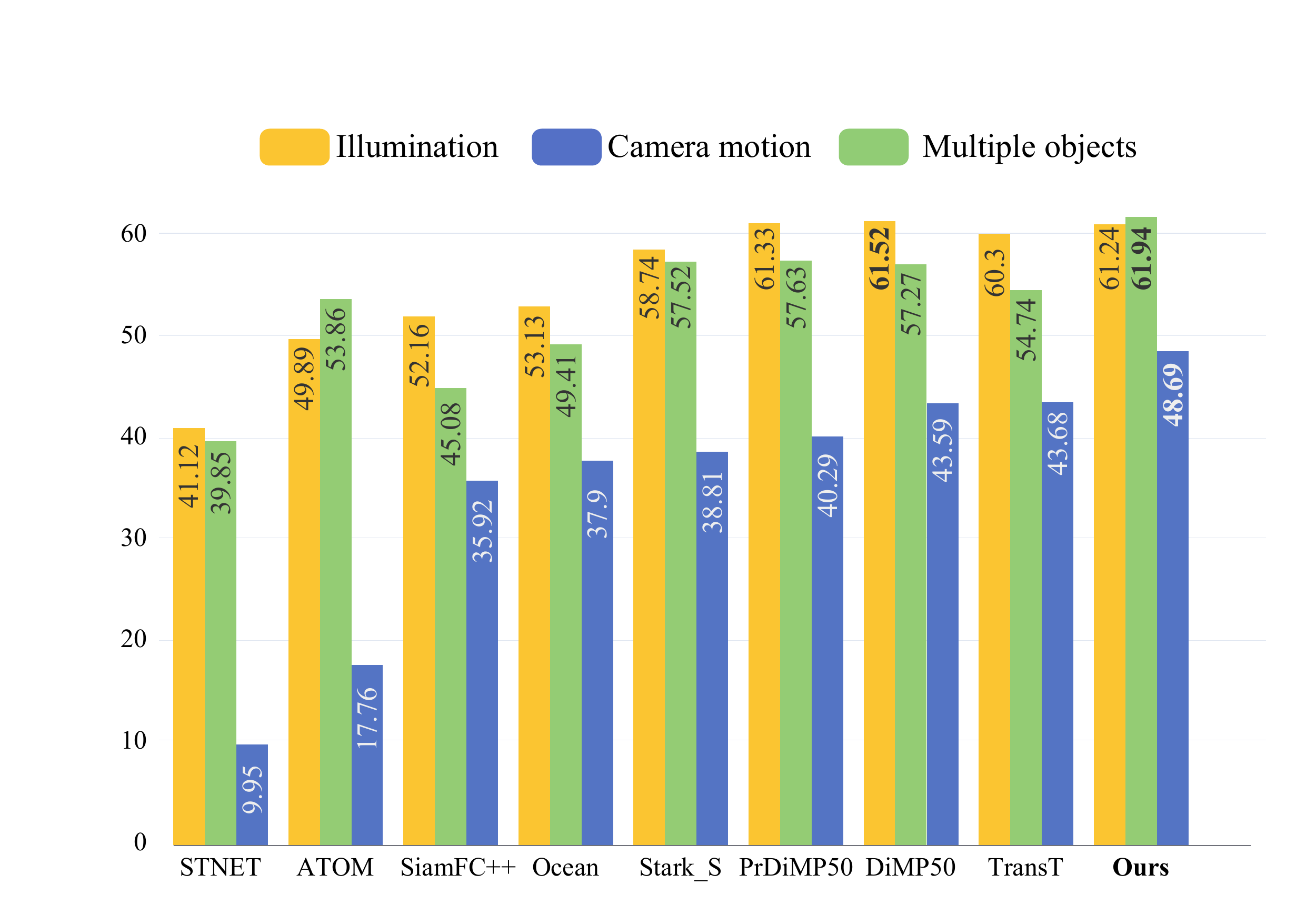}
		
	\end{tabular}}
	\caption{The AUC metric for camera motion sequences,multiple objects and different illumination sequences on FE240 dataset \cite{zhang2021object}. Our tracker leads superior performance in these three challenging scenarios. which evidence that the proposed pipeline is suitable and efficient for event-camera based visual object tracking.}
	\label{fig:scenario}
\end{figure} 
\def\wdenoising{0.9\linewidth}

\begin{figure*}[htb]
    \begin{center}
    \includegraphics[width=\wdenoising]{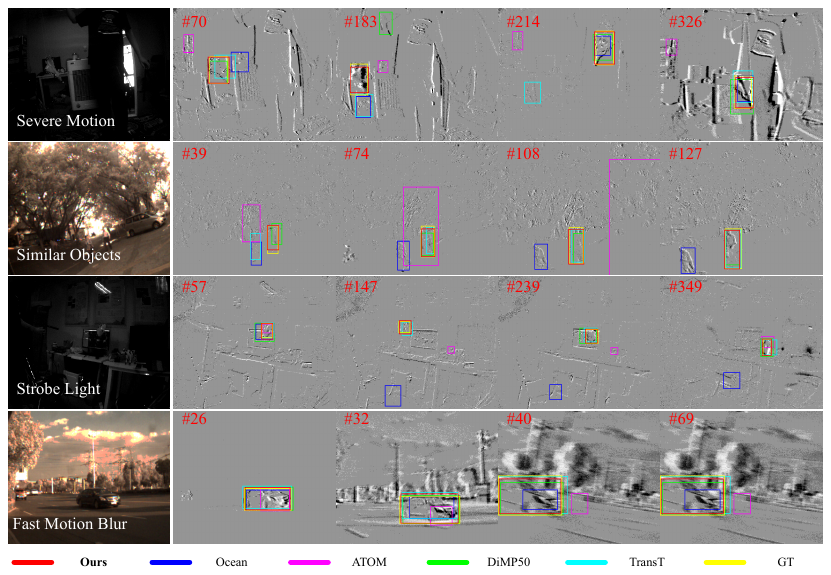}
    \end{center}
    \caption{Four examples to evidence the distractor-aware ability of our proposed tacker against other state-of-the-art trackers on FE240 dataset \cite{zhang2021object} and VisEvent dataset \cite{wang2021visevent}. The first sequence is a severe camera motion sequence, where background noise can severely distract the moving target. The second sequence is a static camera but multiple person are moving together, which becomes a challenge for movement identification. The images in first column are only used for visualization. We can see that the proposed method can still performs well in camera motion scenario and multiple co-moving object scenario. The third and fourth columns are strobe light and fast motion blur scenes, the proposed tracker still exhibits a considerable advantage in terms of performance in both of these scenarios.}
    \label{fig:effect}
\end{figure*}
\setlength{\tabcolsep}{3.8pt}
\begin{table}[t]
	\caption{Loss weight study for the proposed tracker.}
	\centering
	\begin{tabular}{ccc|ccccc}
	\toprule
  & $\lambda_{iou}$ &  $\lambda_{L_1}$ & AUC$\uparrow$ &OP$_{50}\uparrow$ &OP$_{75}\uparrow$ &Prec.$\uparrow$ &Norm Prec.$\uparrow$ \\
	\midrule

	\textit{A.}	& 1 & 2  & 55.49  & 67.75  & 24.11 & 87.54 & 56.16	\\
	\textit{B.} &1  & 1   &  55.97 & 69.92 & 24.77 & 87.50 & 58.06	\\
	\textit{C.}	&2 & 1  & 56.59  & \textbf{70.50} & 24.10 &88.91 & \textbf{58.57} 	\\
	\textit{{Ours}} &1 &5  &  \textbf{56.86} & 69.94 & \textbf{25.03} & \textbf{89.19} & 58.37 	\\
	\bottomrule
	\end{tabular}

	\label{tab:loss}
\end{table}
\paragraph{Comparison on VisEvent Dataset}
Since the VisEvent dataset \cite{wang2021visevent} has many sequences that target objects are similar, static, fully occluded, or completely leave the video sequences, all the trackers suffer from performance degradation. However, our tracker still outperforms other trackers for most evaluation metrics. In particular, the proposed tracker obtains 40.06\%, 55.88\%, and 55.47\% in terms of AUC, Precision, and Normalized Precision, which is described in Table \ref{tab:sota}. Since STNet \cite{zhang2022spiking} needs to split an event frame into 5 parts, we failed to test on the dataset as some raw events are lost. We also report the success plot and normalized precision plot in Figure \ref{fig:eed_plot}. The visualization of tracking similar objects and fast motion blur scenes can be seen on the second and fourth row of Figure \ref{fig:effect}, where our tracker can keep tracking the target person precisely and also demonstrate excellent tracking performance in fast motion blur scenes. More visual examples are shown in Figure \ref{fig:statistics} line 3-4.
\paragraph{Comparison on EED Dataset}
We present the performance comparison of target tracking metrics on the EED dataset in Table \ref{tab:sota}, demonstrating that our proposed DANet achieves favorable performance across multiple metrics. It is noteworthy that the targets in this dataset have smaller sizes and shorter sequence lengths and the sensor used to capture this dataset is older. Furthermore, the data annotation in the dataset is not as accurate as that in the FE240 dataset \cite{zhang2021object} and VisEvent dataset \cite{wang2021visevent}. Nevertheless, these experimental results still demonstrate the tracker's robustness on event cameras of different models. More visualizations are illustrated in Figure \ref{fig:sup_effect} line 5-6.

\paragraph{Loss Weights}
To find the optimal loss weights, as shown in Table \ref{tab:loss}, we try different loss weight combinations for IOU loss and L1 loss. According to the results, the best weight combination for $\lambda_{IOU}$ and $\lambda_{L_1}$ is 1:5, which surpasses the other weight combinations to a great extent.

\paragraph{Efficiency}
It takes about 7 hours to train DANet and the tracker is able to perform inference at 88 FPS on a single V100 GPU. We report the tracking speed and AUC for the top 4 trackers in Table \ref{tab:speed}. Our tracker surpasses other two trackers in both speed and accuracy, which indicates the efficiency of the proposed method.

\begin{table}[t]
	\caption{Tracking speed and accuracy comparison for the top 4 trackers.}
	\centering
	\begin{tabular}{c|cccc}
	\toprule
	Method & TransT \cite{chen2021transformer} & PrDiMP50 \cite{danelljan2020probabilistic} &DiMP50 \cite{bhat2019learning} & DANet    \\
	\midrule
	AUC$(\%)\uparrow$ & 53.46 & 53.43  & 54.49   & 56.86   \\
	\midrule
	FPS $\uparrow$ & 50  &25  &43   & \textbf{88}   \\
	\bottomrule
	\end{tabular}
	\label{tab:speed}
\end{table}

\subsection{Ablation Study}
\begin{table*}[t]
	\caption{quantitative ablation results that demonstrate the effectiveness of our proposed modules on the \textbf{FE240} dataset. \Checkmark means the module is fully used, "SA" denotes the self-attention in each module. "ECA+CFA" is the proposed fusion block used in \cite{chen2021transformer}.TFM(S) denotes add a shortcut into TFM.}
	\centering
	\begin{tabular}{c|ccc|ccccccc}
		\toprule
		& TAN &MAN &Integration & AUC$\uparrow$ &OP$_{50}\uparrow$ &OP$_{75}\uparrow$ &Prec.$\uparrow$ &Norm Prec.$\uparrow$  &Params. &FLOPs \\
		\midrule
    \textit{A.}	& w/o decoder & &N/A  & 55.59 &68.48 &25.29    & 86.52 
	&56.58 &4.30M &3.27G
	\\
	 \textit{B.}	& &\Checkmark &N/A  & 53.00  & 64.12 &19.75 &85.81 & 52.58
	&3.71M &2.95G
	\\
	\midrule

	\textit{C.} &w/o SA &\Checkmark &TFM &55.91 &68.84 &24.36 &87.64 &57.29
	&7.07M &3.61G \\
	\textit{D.} &\Checkmark &w/o SA &TFM &53.40 &66.15 &22.23 &84.39 &54.20 &7.33M &3.65G
	\\
	 \textit{E.} & w/o SA &w/o SA &TFM &55.77 &69.35 &23.54 &87.59 &57.16 &6.93M &3.52G
	\\
	\midrule
	 \textit{F.}	&\Checkmark &  &TFM & 54.96 &67.23 &24.35 &86.26 &55.15 &6.28M
	&3.30G \\
	 \textit{G.}	 &w/o decoder &\Checkmark &ADD &56.53 &\textbf{69.96} &25.14 &88.58 &\textbf{58.60} &5.48M &3.70G    \\
	 \textit{H.} &N/A &N/A &ECA+CFA \cite{chen2021transformer} &55.93 &69.75 &24.06 &87.50 &57.14 &5.75M &3.86G \\
  \textit{I.} & \Checkmark & \Checkmark & TFM &56.32 &69.47 &25.26 &88.12 &57.44  &7.46M &3.73G  \\
	\midrule
	\textit{\textbf{Ours}} &\Checkmark &\Checkmark &TFM(S) &\textbf{56.86} &69.94 &25.03 &\textbf{89.19} &58.37 &7.46M &3.73G        
	\\

	\bottomrule
	\end{tabular}

	\label{tab:ablation}
\end{table*}

We are motivated that event frame can be used to not only distinguish target objects but also help to introduce motion cues to help find moving objects. Thus, our ablation mainly focuses on the following parts: (a) The impact of the proposed TAN and MAN; (b) The benefits of self-attention modules in each proposed network; (c) The efficiency of the fusion mechanism; (d) The impact of different feature extractors. (e) The event representation. The following experiments are carried out on the FE240 dataset \cite{zhang2021object}. All the experiment results in Table~\ref{tab:ablation} have shown the superior performance of our proposed tracking pipeline for event camera.

\paragraph{Module}
As introduced in Section III, the target-aware network aims to match the template frame and the search frame to propose target embeddings, while the motion-aware network aims to extract moving object features. Thus, we evaluate different network structures elaborated below.
(\textit{A}) we remove the decoder of TAN, and the enhanced feature map is then sent to the regression head to predict the final bounding box. We notice the performance degradation by 1.27\%, 2.67\% and 1.79\% in terms of AUC, precision, and normalized precision.(\textit{B}) we remove the TAN, so the information from the template frame does not interact with the search frame. We observe a big performance degradation of 3.86\%. Note that the search frame keeps updating according to the latest tracking results. As MAN only provides object motion information, which degrades our model to be a detector, and thus leads to a significant performance drop. Nonetheless, the tracking performance still outperforms many trackers in Table \ref{tab:sota}.

\paragraph{Importance of self-attention}
{We also tried to find the importance of the self-attention module in transformer block. The experiments below use both the TAN and MAN, the difference is whether the self-attention is applied in each network. (\textit{C}) we remove the self-attention units in TAN; (\textit{D}) we remove the self-attention units in MAN, and (\textit{E}) we remove the self-attention units in both TAN and MAN. Compared with our final configuration, we can see that the self-attention module helps to improve the tracking performance without introducing too many parameters and computational costs.}

\paragraph{Fusion mechanism}
To validate the efficiency of our fusion mechanism, we measure and compare the performance of different structures. The TFM denotes use TFM but removes the shortcut come from TAN, and TFM(S) denotes use the shortcut. (\textit{F}) we remove the MAN and fuse the encoder output and decoder output of TAN, which leads to poor performance. (\textit{G}) we replace the attention-based fusion by concatenating the outputs of TAN encoder and MAN. We observe that the model performance decreases by 0.33\% and 0.61\% in terms of AUC and precision. We also evaluate the fusion mechanism proposed in \cite{chen2021transformer}, which also uses transformer to interact with features from target frame and search frame. (\textit{I}) we use TFM without the shortcut from TAN, thus the features from the encoder of TAN are wasted. We show the superior performance of the tracker according to the results, which indicates the fusion model contributes to the overall performance.

\paragraph{Impact of different feature extractors}
We further investigate the impact of different backbones and we show the comparison results in Table \ref{tab:backbone}, including ResNet \cite{he2016deep} with varied depths, ResNext \cite{xie2017aggregated}, AlexNet \cite{krizhevsky2012imagenet}, and VGG19 \cite{simonyan2014very}. For ResNet \cite{he2016deep}, VGG19 \cite{simonyan2014very} and ResNext \cite{xie2017aggregated}. We prune the later layers of these networks and applied 1x1 convolution before inputting them into TAN and MAN to accommodate the input size requirements of the subsequent Transformer Encoder structure.
According to the results, except for AlexNet \cite{krizhevsky2012imagenet}, the complexity of the feature extractors does not make much difference to the tracker performance but it can significantly influence the model size and the computational cost, so we only use ResNet-18 \cite{he2016deep} as our final feature extractor, which has the lowest parameters and computation costs but still archive the best performance in AUC, precision and normalized precision, respectively.
\begin{table}[t]

	\caption{Comparison for the proposed tracker with different feature extractors.}
	\centering
	\footnotesize
	\setlength{\tabcolsep}{4.4pt}

	\begin{tabular}{l|ccccc}
	\toprule
  Backbones & AUC$\uparrow$ &Prec. $\uparrow$ &Norm Prec. $\uparrow$  & Params.  & FLOPS\\
	\midrule

	  AlexNet \cite{krizhevsky2012imagenet}   & 48.38 & 76.64 & 48.91 & 14.82M & 10.49G \\
    VGG19  \cite{simonyan2014very}   & 56.02 & 87.38 & 57.01 & 25.19M & 43.70G \\
    ResNet-34 \cite{he2016deep} & 56.85 & 88.42 & 59.22 & 8.13M  & 5.24G   \\
	  ResNet-50 \cite{he2016deep} & 55.43 & 87.41 & 56.82 & 18.81M & 11.20G	\\
	  ResNext-50 \cite{xie2017aggregated} & 55.73 & 87.20 & 58.07 & 18.78M & 11.27G \\
	  ResNet-18 \cite{he2016deep} &\textbf{56.86} & \textbf{89.19} &\textbf{59.37} 
	& \textbf{7.46M} & \textbf{3.73G} \\
	\bottomrule
	\end{tabular}

	\label{tab:backbone}
\end{table}

\paragraph{Impact of event representation}
We carry several experiments to find suitable event representations. On the one hand, For a fixed time interval, we use event image \cite{Rebecq2017RealtimeVO} and voxel grid \cite{zhu2019unsupervised} to form the input of the proposed tracker, the results is shown in Table \ref{tab:represent}, we can see the used event representation performs best across most metrics. On the other hand, since the FE240 dataset provide up to 240Hz annotation, we can use different time windows with the used event representation. The results can be seem in Figure \ref{fig:frame-rate}, which indicates that 40Hz works best for the proposed tracker. We notice a performance drop with the increments on event frame rate, we believe this is due to the fact that the automatic annotation of the FE240 dataset \cite{zhang2021object} exhibits lower accuracy as the frame rate increases. Additionally, the event data used as input to the network becomes sparser at higher frame rates, which also affects the performance of the tracker.
\begin{table}[t]
	\caption{Event representation study for the proposed tracker.}
	\centering
	\begin{tabular}{l|ccccc}
	\toprule
  Representation  & AUC$\uparrow$ &OP$_{50}\uparrow$ &OP$_{75}\uparrow$ &Prec.$\uparrow$ &Norm Prec.$\uparrow$ \\
	\midrule
    Event Frame~\cite{Rebecq2017RealtimeVO} 	 & 54.20 & 66.74 & 22.11 & 86.12 & 54.80 \\
    Voxel Grid~\cite{zhu2019unsupervised}   & 55.94 &68.22 &\textbf{25.72} & 87.54 & 55.70 \\
      Ours	 & \textbf{56.86} &\textbf{69.94} & 25.03 & \textbf{89.19} & \textbf{58.37} \\
	\bottomrule
	\end{tabular}
	\label{tab:represent}
\end{table}

\def\wdenoising{0.9\linewidth}
\def\signw{0.63\linewidth}
\begin{figure}[htbp]
	\centering
	
	\begin{tabular}{c}
	
	\includegraphics[width=\wdenoising]{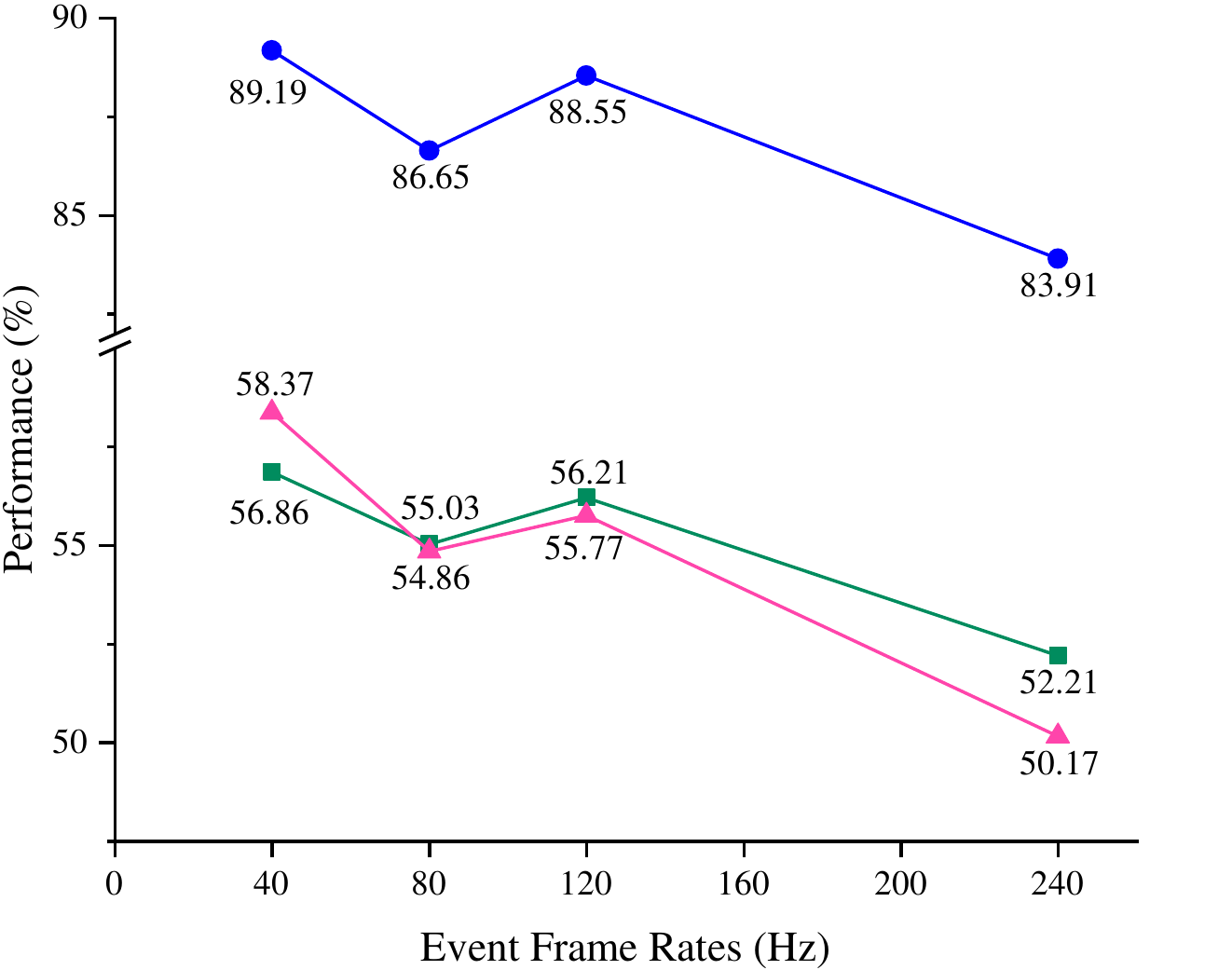} \\
    \includegraphics[width=\signw]{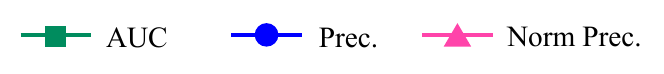} \\
		
	\end{tabular}
	\caption{Performance of various event frame rates on our tracker.}
	\label{fig:frame-rate}
\end{figure} 

\subsection{Feature visualization}
Figure \ref{fig:feature_map} shows an example of feature maps from FE240 dataset \cite{zhang2021object}. The images from left to right are grayscale image, the cropped and padded event frame, the output feature maps w/o and w/ MAN for the last convolution layer of the center regression branch. We can observe that using the MAN improves object localization.

\def\wdenoising{1\linewidth}
\begin{figure}[htb]
	\setlength{\tabcolsep}{0.8pt}
	\centering
	
	\begin{tabular}{cc}
	
	\includegraphics[width=\wdenoising]{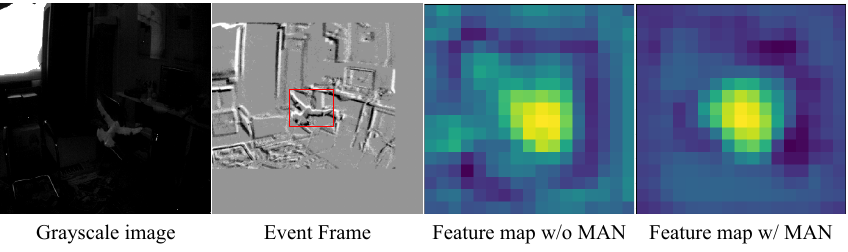}
		
	\end{tabular}
	\caption{The feature map output from the last convolution blocks. The crop and padded event search frame and the target are visualized in the second column.}
	\label{fig:feature_map}
\end{figure} 
\subsection{Fuse event and frame for robuster tracking}
Since FE240 dataset provides aligned event stream and grayscale frame, we can jointly utilize the information from event domain and RGB domain. In this experiment, we use add operator to fuse features and design several fusion strategies based on the proposed pipeline. The fusion operation is performed after the Feature Extract Network. Table \ref{tab:fusion} shows the results for different fusion methods. As the RGB frames suffer from HDR and fast-moving scenes, the performance of a tracker relying exclusively on RGB images is significantly compromised. We further investigate that employing RGB features as inputs for the Target-Aware Network (TAN) and integrating RGB features with event features as inputs for the Motion-Aware Network (MAN) effectively enhances the tracking performance through multimodal fusion. Compared to event only tracker, this fusion approach results in improvements of 1.01\%, 0.14\%, and 3.20\% in AUC, Precision, and Normalized Precision metrics, respectively. We attribute this improvement to the rich target texture information present in RGB images, which significantly improves the target discrimination capability of the TAN network.
\begin{table}[t]
	\caption{Study for fusing RGB domain and event domain based on the proposed tracker.}
	\centering
	\begin{tabular}{ccccc}
	\toprule
  TAN & MAN &  AUC$\uparrow$ &Prec.$\uparrow$ &Norm Prec.$\uparrow$ \\
	\midrule

	Frame & Frame   & 38.53 & 64.28 & 34.44 \\
	Event & Event   & 56.86 & 89.19 & 58.37 \\
    Frame + Event & Frame + Event   & 56.24 & 86.98 & 57.82 \\
    Frame + Event & Event   & 57.32 & \textbf{90.74} &57.55 \\

    Frame & Frame + Event  & \textbf{57.87} & 89.33 & \textbf{61.57} \\

	\bottomrule
	\end{tabular}

	\label{tab:fusion}
\end{table}

\subsection{Discussion}
\paragraph{Limitation}
The performance of the proposed offline tracker primarily relies on the quality of the search event frames. As shown in Figure \ref{fig:fail}, when there is adequate events of target, the tracker can effectively locate the target. However, in cases where the events of the tracked object are sparse, target discrimination extract from TAN and motion feature extract from MAN are unreliable, leading to tracking failures and potential degradation of subsequent tracking performance. One possible approach is to assign a confidence score to the objects in the search event frames and multiply this score with the regression box. In this way, when the score is high, the tracking result can be given high confidence, while for low scores, the output tracking box will be restricted around the previous location.

\def\wdenoising{1\linewidth}
\begin{figure}[htb]
	\setlength{\tabcolsep}{0.8pt}
	\centering
	
	\begin{tabular}{cc}
	
	\includegraphics[width=\wdenoising]{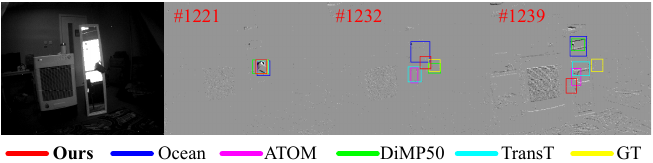}
		
	\end{tabular}
	\caption{Illustration of failure cases. Since the object generate very few events in the search frame, there is not enough object features computed in the proposed network, as a result, the tracker can not identify the target in these frames and outputs imprecise bounding boxes.}
	\label{fig:fail}
\end{figure} 
\paragraph{Future work}
The proposed tracking method is designed for event-camera based visual object tracking. However, we do not leverage the asynchronous natural for event camera. Inspired by \cite{sekikawa2019eventnet}, a promising future research direction is to treat the raw input events as 3D point cloud and use PointNet++ \cite{qi2017pointnet++} based method or graph-based method such as \cite{zhu2022learning} to extract spatial-temporal features. As event stream can be treated to spike train, another strategy is to use Spiking Neural Network (SNN) \cite{wu2018spatio} to process event data. Several works have managed to use Conv-SNN to process and propagate the event frame for vision tasks  \cite{lee2020spike,gehrig2020event,duwek2021image}, and there is still a promising direction on training the pure Spiking Neural Network for visual tracking with the asynchronous events as input.

\section{Conclusion}

In this paper, we proposed an event-based tracker that is based on a Siamese network structure and introduces a motion-aware network and target-aware network to exploit both motion cues and edge cues from the event camera, and use Target-motion Fusion module to fuse the features for regressing the target coordinates. We used two public event-based tracking dataset to validate the efficiency of our proposed tracker, the extensive validation and ablation experiences showed that our work has a significant performance promotion, and the inference speed is much faster than other online trackers. Our future work will focus on modeling the motion pattern for target movements and making use of the asynchronous feature of event data.

\section*{Acknowledgments}
This work was supported in part by National Key Research and Development Program of China (2022ZD0210500), the National Natural Science Foundation of China under Grants  61972067/U21A20491, the Distinguished Young Scholars Funding of Dalian (No. 2022RJ01) and the HiSilicon(Shanghai) Technologies Co.,Ltd (No.TC20210510004).

\bibliographystyle{IEEEtran}
\bibliography{paper_bib}

\newpage

 




\vfill

\end{document}